\documentclass[10pt,twocolumn, letterpaper]{article}
\usepackage[pagenumbers]{iccv}
\usepackage{amsfonts}  
\usepackage{titlesec}
\titleformat{\section}{\normalfont\Large\bfseries}{\thesection}{1em}{}

\title{Winning the MIDST Challenge: New Membership Inference Attacks on Diffusion Models for Tabular Data Synthesis}

\author{Xiaoyu Wu, Yifei Pang, Terrance Liu, Steven Wu \\ 
Carnegie Mellon University\\
Pittsburgh, PA 15213, USA \\
{\tt\small nicholaswu2022@gmail.com, \{yifeip, terrancl\}@andrew.cmu.edu, zstevenwu@cmu.edu}}

\begin{document}

\maketitle
\begin{abstract}
Tabular data synthesis using diffusion models has gained significant attention for its potential to balance data utility and privacy. However, existing privacy evaluations often rely on heuristic metrics or weak membership inference attacks (MIA), leaving privacy risks inadequately assessed. In this work, we conduct a rigorous MIA study on diffusion-based tabular synthesis, revealing that state-of-the-art attacks designed for image models fail in this setting. We identify noise initialization as a key factor influencing attack efficacy and propose a machine-learning-driven approach that leverages loss features across different noises and time steps. Our method, implemented with a lightweight MLP, effectively learns membership signals, eliminating the need for manual optimization. Experimental results from the MIDST Challenge @ SaTML 2025 demonstrate the effectiveness of our approach, securing first place across all tracks. Code is available at \url{https://github.com/Nicholas0228/Tartan_Federer_MIDST}.
\end{abstract}

\iccvsection{Introduction}

Tabular data synthesis has garnered significant attention for its potential to safeguard data privacy~\cite{tab1, tab2, tab3, tab4, tab5}, providing a means to facilitate downstream tasks while possibly mitigating the risk of user information leakage. Recent studies have explored the adaptation of advanced diffusion models for tabular data synthesis, claiming that these models achieve significantly higher generation quality while maintaining privacy~\cite{tab3, tab4, tab5}.

Although the assertion that synthetic data enhances privacy is widely accepted in the literature, it has not been rigorously verified in many scenarios. For instance, prior works on diffusion-based tabular data synthesis have either primarily relied on basic privacy evaluation methods, such as Distance to Closest Records (DCR)~\cite{tab4, tab5}, or measure under membership inference attack (MIA) but using some basic attacking method~\cite{tab3}. These approaches serve as intuitive measurements but may underestimate the extent of potential privacy leakage.

In this paper, we employ MIA to assess the privacy risks of diffusion models for tabular data, ensuring a broad coverage of attack methods to provide a comprehensive evaluation. MIA~\cite{shokri2017membership} offers a direct and informative measure of privacy leakage by explicitly quantifying an attacker's ability to distinguish training data from non-training data. However, we show that existing MIA techniques designed for diffusion models are not directly applicable to tabular data synthesis. This necessitates a more rigorous evaluation, and we demonstrate that significantly stronger attacks can be developed for this setting.

Specifically, we first apply SecMI~\cite{duan2023diffusion}, one of the strongest established attacks for image-based diffusion models. However, we find that its performance on tabular data diffusion models is nearly indistinguishable from random guessing. This highlights the fundamental differences in data distribution and structure, suggesting that tabular data requires specialized attack methods tailored to its unique characteristics.

Further investigation reveals that the choice of initialized noise significantly impacts attack performance, leading to high variance across different noise selections. A natural approach to addressing this issue is to identify an optimal noise configuration. However, this is non-trivial, as prior studies on diffusion models suggest that effective noise selection often requires complex optimization~\cite{ma2024could} or additional training data~\cite{zhou2024golden}. To circumvent this challenge, we propose a more direct, machine-learning-driven approach that mitigates the influence of noise selection and the time variable \( t \).  

Specifically, we treat loss values as features, computing multiple losses for each data point using different noises and time variables. These loss features are then processed by a lightweight three-layer MLP to predict membership status. Intuitively, this approach enables the model to automatically learn the relationship between noise, time, and membership information, eliminating the need for manual optimization or heuristic selection.  

Experimental results from the MIDST Challenge @ SaTML 2025~\cite{midst2025} demonstrate the effectiveness of our method, securing first place across all tracks.  

\iccvsection{Related works}
\iccvsubsection{Diffusion Models for Tabular Data Generation}

Diffusion models have been increasingly adopted for tabular data generation, with various approaches designed to handle its unique structure and constraints. TabDDPM~\cite{tab3} employs denoising diffusion models with separate diffusion processes for numerical and categorical data. TabSyn~\cite{tab4} integrates a transformer-based variational autoencoder with latent diffusion modeling. ClavaDDPM~\cite{tab5} leverages clustering labels as intermediaries to capture relationships between tables, particularly enforcing foreign key constraints, thereby enabling multi-table tabular data generation.

\iccvsubsection{Membership Inference Attack on Diffusion Models}

Diffusion models (DMs)~\citep{ho2020denoising,sohl2015deep} generate data by iteratively denoising a Gaussian-sampled variable. They consist of a forward process, where noise \(\varepsilon \sim \mathcal{N}(0,1)\) is progressively added to data \(x_0\), forming noisy samples:
$
x_t = \sqrt{\alpha_t} x_0 + \sqrt{1 - \alpha_t} \varepsilon.
$

In the backward process, a model \(\epsilon_{\theta}\) predicts and removes noise from \(x_t\). The training objective, is known as the diffusion loss:
$
\mathcal{L}_{DM} = \mathbb{E}_{\varepsilon\sim \mathcal{N}(0,1), t}\left[\|\epsilon_{\theta}(x_t, t) - \varepsilon\|_2^2\right].
$
This loss guides the model to reconstruct clean data from noisy inputs, enabling effective data generation.

MIA on diffusion models varies in design, primarily focusing on deriving strong loss functions to better assess memorization. Among them, SecMI~\cite{duan2023diffusion} is one of the most influential approaches. It leverages the DDIM-forwarding process to actively add noise to a given sample, represented as \( x_{t+1} = \phi_{\theta}(x_t, t) \), and the general backward process to perform one-step denoising, denoted as \( \psi_{\theta}(x_t, t) \).  Membership is then determined using a \( t \)-error metric:  

\begin{equation}
       \tilde{l}_{t,x_0} = \|\psi (\phi (\tilde{x}_t, t), t) - \tilde{x_t} \|_2, 
\end{equation}
where  $ \tilde{x}_t = \Phi_{\theta}(x_0, t)=\phi_\theta(\dots \phi_\theta(\phi_\theta(x_0, 0), 1), t-1).$

\iccvsection{Threat Model and Evaluation}  

We consider a MIA against a diffusion model trained for tabular data synthesis. Let the target diffusion model be parameterized by $\theta$ and denoted as $\epsilon_{\theta}$. Given a dataset $\mathcal{D} = \{x_i\}_{i=1}^{N}$, the model is trained to approximate the underlying data distribution and generate synthetic samples.

\paragraph{Adversary's Goal.}  
The adversary aims to determine whether a given data point $x$ was part of the training dataset $\mathcal{D}$. Specifically, given a query sample $x$, the adversary must infer the binary membership status $m_x \in \{0,1\}$, where $m_x = 1$ if $x \in \mathcal{D}$ and $m_x = 0$ otherwise.

\paragraph{Adversary's Knowledge.}  
We consider two attack settings: black-box and white-box. The setup consists of:  
30 train-phase models \(\{\epsilon_{\theta^{(i)}}\}_{i=1}^{30}\), where the adversary has access to their training samples.  
 20 dev-phase models \(\{\epsilon_{\theta^{(i)}}\}_{i=31}^{50}\) and 20 final-phase models \(\{\epsilon_{\theta^{(i)}}\}_{i=51}^{70}\), for which training samples are not accessible.

In the white-box setting, the adversary has access to:

\begin{itemize}
    \item The full model weights of the 30 train-phase models \(\{\epsilon_{\theta^{(i)}}\}_{i=1}^{30}\), along with their corresponding training samples.  
    \item The model weights of the 20 dev-phase and 20 final-phase models \(\{\epsilon_{\theta^{(i)}}\}_{i=31}^{70}\), but without access to their training samples.  
\end{itemize}  

With this access, the adversary attempts to infer the membership status of a given sample \(x\) using an adversarial function \(A(\cdot)\), which directly takes the model as input to compute a membership prediction:

\begin{equation}
     A(\epsilon_{\theta^{(i)}}, x) \rightarrow \hat{m}_x.   
\end{equation}

In the black-box setting, the adversary has access to:

\begin{itemize}
    \item The training samples of the 30 train-phase models, but not the models themselves.  
    \item A synthetic dataset \(\{X_g^{(i)}\}_{i=1}^{70}\), containing an equal number of samples generated by the given models across all phases, without access to their weights.
\end{itemize}  

In this case, the adversary relies on model-generated samples to make a membership prediction:

\begin{equation}
    A(X_g^{(j)}, x, \{X_{g^{(i)}}\}^{30}_{i=1}) \rightarrow \hat{m}_x.
\end{equation}

For each model \(\epsilon_{\theta^{(j)}}\), the adversary is provided with a balanced dataset consisting of an equal number of member and holdout samples.

\paragraph{Evaluation Metrics.}  
By default, we use Area under the curve (AUC) as the primary evaluation metric. Following the MIDST competition~\cite{midst2025}, we also evaluate the attack using the true positive rate (TPR) at a fixed low false positive rate (FPR).

\iccvsection{Proposed Methodology}

We have developed a structured and adaptive approach for conducting MIA on tabular diffusion models. Our method comprises three key steps: model preparation, loss function selection, and loss processing using an MLP model. The following sections will provide a detailed discussion of each step.

\iccvsubsection{Model Preparation}

Since our MIA relies on loss values, the first step is to prepare a diffusion model associated with the training data. This is the foundation for subsequent processes.

For model preparation, we consider both white-box and black-box settings:
\begin{itemize}
    \item \textbf{White-box scenario}: The attack directly utilizes the provided models without modification, with full access to internal parameters.
    \item \textbf{Black-box scenario}: As the target model is inaccessible, we train a shadow model using its synthetic data under the same configuration. However, instead of explicitly simulating the target model, the shadow model serves to extract additional information of the training data. 
\end{itemize}

\iccvsubsection{Loss Function Selection}

An effective loss function plays a crucial role in distinguishing between training and holdout samples. To maximize their loss difference, we explored four different loss functions. Among them, three consistently performed well in tabular MIA tasks.  

\begin{figure}
    \centering
    \includegraphics[width=0.8\linewidth]{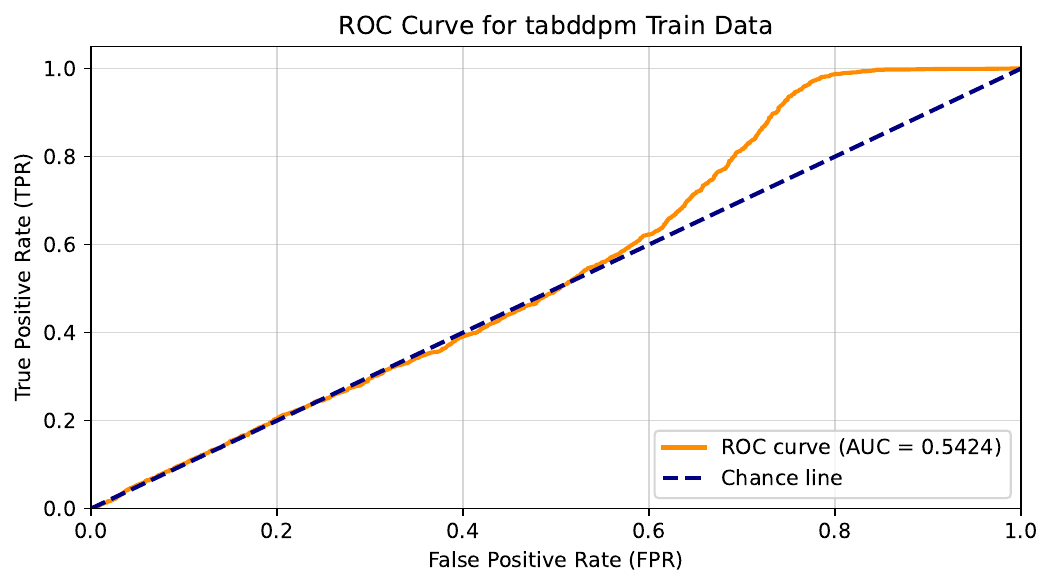}
    \caption{ROC of SecMI in white box TabDDPM MIA}
    \label{fig:SecMI}
\end{figure}

We start with a baseline evaluation using SecMI~\cite{duan2023diffusion}, a widely used and effective method for diffusion models. However, as shown in Fig.~\ref{fig:SecMI}, our experimental results reveal that SecMI~\cite{duan2023diffusion} performs close to random guessing, indicating its limited effectiveness in this setting.

Contrary to the findings in~\cite{duan2023diffusion}, we observe that a naive approach—fixing a randomly initialized noise \(\varepsilon_0\) and a fixed time variable \( t \)—achieves significantly better performance in membership inference. Specifically, this approach uses the direct loss function:

\begin{equation}
    l(\epsilon_{\theta}, x_0, \varepsilon_0, t) = \left\|\epsilon_{\theta} \left( \sqrt{\alpha_t} x_0 + \sqrt{1 - \alpha_t} \varepsilon_0, t \right) - \varepsilon_0 \right\|_2^2.
\end{equation}

Membership is then determined by evaluating:

\begin{equation}
   A(l(\epsilon_{\theta^{(i)}}, x, \varepsilon_0, t), x). 
\end{equation}

This simple yet effective method significantly improves attack performance compared to SecMI~\cite{duan2023diffusion}. We refer to this method as ``Naive Loss."

\begin{figure}[t]
  \begin{subfigure}{0.49\linewidth}
\includegraphics[width=\linewidth]{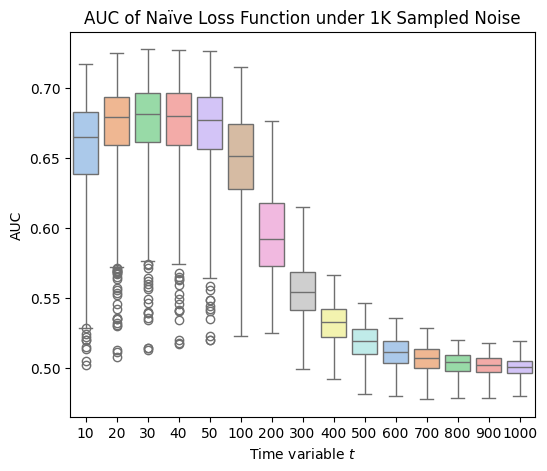}
    \caption{TabDDPM}
  \end{subfigure}
        \begin{subfigure}{0.49\linewidth}
    \includegraphics[width=\linewidth]{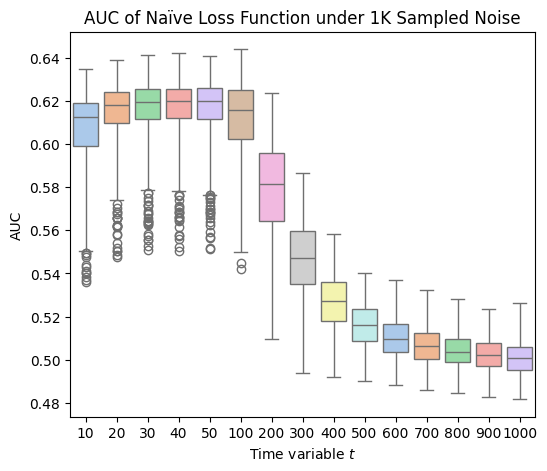}
    \caption{ClavaDDPM}
  \end{subfigure}
\caption{AUC for 1,000 different fixed noise in MIA targeting TabDDPM and ClavaDDPM, exhibiting high variance, ranging from random guessing to partially successful attacks.}
      \label{variance2}
\end{figure}

However, despite its improved effectiveness, this Naive Loss method suffers from high variance. As illustrated in Fig. \ref{variance2}, if we sample 1,000 different noise and select the best-performing one for each model, we can achieve substantially stronger results. Nevertheless, this is impractical in real-world attack scenarios, as an adversary would not have the capability to systematically explore and select the optimal noise configuration. Therefore, there comes a question about how to process these losses for a stable and accurate membership inference.


\iccvsubsection{Loss Processing and Membership Inference}
Since raw loss values vary due to different noises and time variables \( t \), a simple loss processing approach (e.g. threshold-based splitting) is insufficient for robust inference. To address this challenge, we propose a machine-learning-driven approach. Specifically, we introduce a three-layer Multi-Layer Perceptron (MLP) to model the relationship between loss values and membership status, improving attack accuracy.


\begin{table*}[t]
    \centering
    \caption{Selected hyper-parameters configurations.}
    \label{tab:hyper-parameters}
    \resizebox{\textwidth}{!}{%
    \begin{tabular}{|p{2.5cm}|p{3.5cm}|p{2.5cm}|p{3cm}|p{2.5cm}|p{2.5cm}|}
        \hline
        Model & Time Variable Set & Noise Number & Training Samples (per model) & Test Samples (per model) & Training Epochs \\
        \hline
        TabDDPM & [5,10,20,30,40,50,100] & 300 & 6000 & 2,000 & 5000 \\
      \hline
       ClavaDDPM & [5,10,20,30,40,50,100] & 500 & 800 & 400 & 750 \\
        \hline
    \end{tabular}
    }

\end{table*}

\begin{table*}[t]
\centering
\caption{Comparison of our methods with the baseline under white-box scenarios, with all results measured by AUC and TPR@10\%FPR. Best Noise refers to the Naive Loss method with the best noise selected from 1,000 different instances.}
\label{comparision}
\begin{tabular}{ccccc}
\hline
 & \multicolumn{2}{c}{TabDDPM}& \multicolumn{2}{c}{ClavaDDPM} \\
 & AUC & TPR@10\%FPR & AUC &  TPR@10\%FPR \\ \hline
Naive Loss ($t=10$) &   0.654   & 0.213 &0.606&  0.206 \\
Naive Loss  ($t=20$) &      0.670 &     0.242  &0.614& 0.221   \\
Naive Loss  ($t=50$) & 0.670      &      0.247      &0.617&   0.221  \\
Naive Loss  ($t=100$)& 0.648      &     0.225  &0.612 & 0.207  \\
Naive Loss  ($t=200$)&    0.595   &   0.172    &0.579 & 0.168  \\
Naive Loss  ($t=1000$)      &   0.500    &      0.102 & 0.500 & 0.104  \\ \hline

Best Noise ($t=10$)  &  0.717&0.337&     0.634 &      0.279\\ 
Best Noise ($t=20$)    &  0.724&0.353&    0.639  &    0.286  \\ 
Best Noise ($t=50$)  &  0.726&0.363&    0.641  &  0.289    \\ 
Best Noise  ($t=100$) &  0.715&0.343&   0.644   & 0.288

\\ \hline

Trained MLP (Our Method)    & \textbf{0.787}      & \textbf{0.448}     &\textbf{0.720} &  \textbf{0.322}   \\ 
\hline
\end{tabular}

\end{table*}

\iccvsubsubsection{Data Preparation}
To prevent overfitting in the MLP model and effectively monitor its training progress, we adopt a model-based train-validation structure. This model-based split helps reduce overfitting, as the validation set remains entirely independent of the training set. Specifically, we:
\begin{itemize}
    \item Split 30 models from the train phase into 20 for training and 10 for validation. For each set, we randomly sample a fixed number of training and holdout data, while ensuring that no overlapping data (including challenge data) is used in MLP training.
    \item Evaluate performance on the challenge dataset, providing an approximate estimate of the final MLP model performance.
\end{itemize}

Next, we generate features by computing loss values under varying noise levels and time variables \( t \) for each data.
We first construct a fixed noise set containing \( n_{\epsilon} \) different noise samples and a fixed time-variable set with \( n_{t} \) distinct \( t \) values. 
Then, for each model and each input, we compute \( n_{\epsilon} \times n_{t}\) loss values, forming a structured dataset \((X, Y)\), where \( X \) represents the loss features and \( Y \) denotes the corresponding membership labels for training and validation.

\iccvsubsubsection{Model Training}
Once the data is prepared, we train a small three-layer MLP to learn the relationship between loss features and membership. The best hyper-parameters are selected based on validation TPR performance for generalization.

\iccvsubsubsection{Membership Inference}
After obtaining the final trained MLP model, we follow the same steps as before to generate prediction scores for challenge data. Specifically, we first extract features using the same \( n_{\epsilon} \) noise samples and \( n_{t} \) distinct \( t \) values. These features are then fed into the model to produce prediction scores. Finally, the scores are clipped to the range \([0,1]\), indicating the confidence score of membership.


\iccvsection{Experiments}

We conducted extensive experiments and hyperparameter tuning across four tasks, identifying a set of hyperparameters that consistently achieves near-optimal performance for each task, as shown in Table \ref{tab:hyper-parameters}. The experimental results are presented in Table \ref{comparision}, demonstrate that our neural network-based approach significantly improves performance.


This method achieved 1st place in the SaTML MIDST competition, ranking 1st in the dev-phase with TPR@10\%FPR scores of 0.25 in the Black-Box Single-Table Competition, 0.23 in the Black-Box Multi-Table Competition, 0.36 in the White-Box Single-Table Competition, and 0.45 in the White-Box Multi-Table Competition. While the final-phase scores are not visible, the leaderboard confirms that our method still secured 1st place across all categories.

\iccvsection{Conclusion}
In this paper, we investigate the privacy risks of diffusion-based tabular data synthesis through a rigorous MIA analysis. We demonstrate that existing MIA techniques designed for diffusion models in image domains are ineffective for tabular data and identify noise initialization as a key factor influencing attack success. To address this, we propose a machine-learning-driven approach that leverages loss features across different noises and time steps, enabling more robust membership inference without manual optimization. Our method significantly outperforms prior attacks, as evidenced by its first-place performance in the MIDST Challenge @ SaTML 2025, highlighting the need for stronger privacy assessments in synthetic tabular data generation.
\newpage
\bibliographystyle{abbrv}
\bibliography{mybib}

\end{document}